\documentclass[lettersize,journal]{IEEEtran}
\usepackage{amsmath,amsfonts}
\usepackage{algorithmic}
\usepackage{algorithm}
\usepackage{array}
\usepackage[caption=false,font=normalsize,labelfont=sf,textfont=sf]{subfig}
\usepackage{textcomp}
\usepackage{stfloats}
\usepackage{url}
\usepackage{verbatim}
\usepackage{graphicx}
\usepackage{cite}
\usepackage{hyperref}
\usepackage{bbding}
\usepackage{amssymb} 
\usepackage{multirow} 

\begin{document}

\title{TIDE: Test Time Few Shot Object Detection}

\author{Weikai Li, Hongfeng Wei, Yanlai Wu, Jie Yang, Yudi Ruan, Yuan Li$^*$ and Ying Tang$^*$,~\IEEEmembership{Senior Member,~IEEE}

\thanks{Yuan Li and Ying Tang are corresponding authors.}
\thanks{Weikai Li is with the School of Mathematics and Statistics, Chongqing Jiaotong University, Chongqing, 400074, China, MIIT Key Laboratory of Pattern Analysis and Machine Intelligence, Nanjing, 210016, China and with  China Science IntelliCloud Technology, Co., Ltd., Anhui, 230000, China  (e-mail: leeweikai@outlook.com).}
\thanks{Hongwei Wei is with the  China Science IntelliCloud Technology, Co., Ltd., Anhui, 230000, China  (e-mail: jason@intellicloud.ai).}
\thanks{Yanlai Wu, Jie Yang and Yudi Ruan are with the School of Information Science and Technology, Chongqing Jiaotong University, Chongqing, 400074, China  (e-mail: \{632007100121, 632007030617 and 632207100104\}@mails.cqjtu.edu.cn). }
\thanks{Yuan Li is with China Science IntelliCloud Technology, Co., Ltd., Anhui, 230000, China (e-mail: lyn@intellicloud.ai). }
\thanks{Ying Tang is with the Department of Electrical and Computer Engineering, Rowan University,  Glassboro, NJ 08028 USA (e-mail: tang@rowan.edu). }
\thanks{Manuscript received April 19, 2021; revised August 16, 2021.}}

\markboth{Journal of \LaTeX\ Class Files,~Vol.~14, No.~8, August~2021}%
{Shell \MakeLowercase{\textit{et al.}}: A Sample Article Using IEEEtran.cls for IEEE Journals}


\maketitle

\begin{abstract}
Few-shot object detection (FSOD) aims to extract semantic knowledge from limited object instances of novel categories within a target domain. Recent advances in FSOD focus on fine-tuning the base model based on a few objects via meta-learning or data augmentation. Despite their success, the majority of them are grounded with parametric readjustment to generalize on novel objects, which face considerable challenges in Industry 5.0, such as (i) a certain amount of fine-tuning time is required, and (ii) the parameters of the constructed model being unavailable due to the privilege protection, making the fine-tuning fail. Such constraints naturally limit its application in scenarios with real-time configuration requirements or within black-box settings. To tackle the challenges mentioned above, we formalize a novel FSOD task, referred to as Test TIme Few Shot DEtection (TIDE), where the model is un-tuned in the configuration procedure. To that end, we introduce an asymmetric architecture for learning a support-instance-guided dynamic category classifier. Further, a cross-attention module and a multi-scale resizer are provided to enhance the model performance. Experimental results on multiple few-shot object detection platforms reveal that the proposed TIDE significantly outperforms existing contemporary methods. The implementation codes are available at \href{https://github.com/deku-0621/TIDE}{https://github.com/deku-0621/TIDE}.
\end{abstract}

\section{Introduction}
\IEEEPARstart{O}{bject} detection is a fundamental computer vision task that deals with detecting instances of visual objects for a certain class (such as humans, animals, or cars),  which has gained much attention in several downstream tasks, e.g.,  autonomous driving \cite{mao20233d}, face recognition \cite{schneiderman2000statistical}, fire detection \cite{muhammad2018efficient} and environmental protection \cite{wu2019intelligent,kong2020iwscr}.  The focus of object detection is to provide one of the fundamental pieces of knowledge required by computer vision applications: the identification and localization of objects. Over the past decade, substantial progress has been made in object detection utilizing deep learning techniques \cite{zou2023object,zhao2019object,ji2021task,zhang2021partial,minaeian2015vision,liu2022deep}. However, these deep learning-based approaches typically demand a considerable amount of training data to extract robust concepts.

\begin{figure}[t]
\centering
\includegraphics[width=0.5\textwidth]{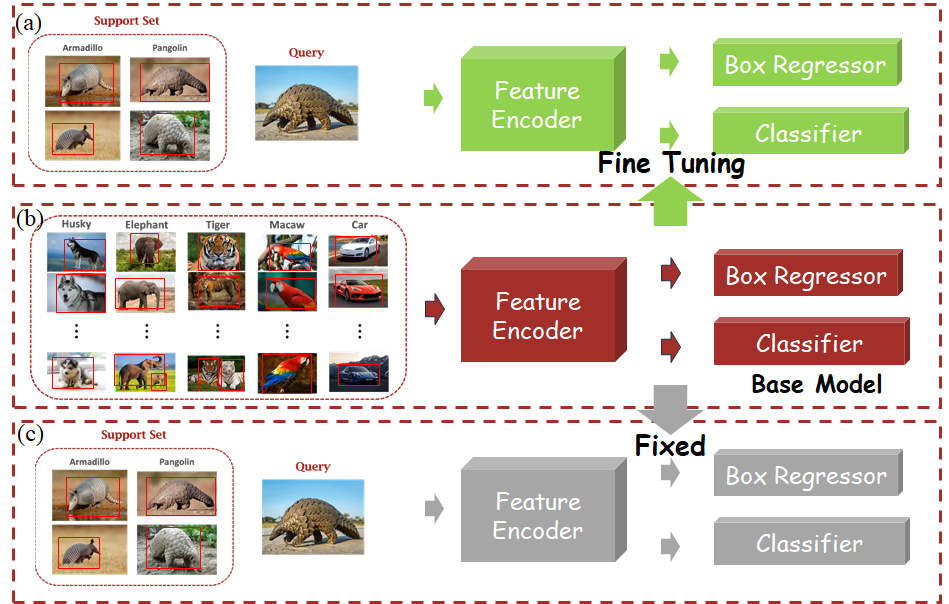}
\caption{FSOD vs TIDE (a) The conventional FSOD approaches to leverage the novel object instance to fine-tune the base model,(b) The base model is initialized by abundant instance-level data, and (c) our designed TIDE approach \textbf{DOES NOT} require fine-tuning the model, the base model can even be a black-box.}
\label{figure1}
\end{figure}

Unfortunately, generating a large number of annotated images for object recognition is both time-consuming and resource-intensive. In certain contexts, such as medical applications \cite{katzmann2021explaining}, the identification of endangered species \cite{mannocci2022leveraging}, or the industry application \cite{huang2023industrial}, accumulating a wealth of images is even unattainable. In contrast, unlike traditional deep learning approaches, human beings can easily recognize new objects and grasp the base concept of a category with minimal data even in their early years \cite{smith2002object,samuelson2005they}. When children encounter novel objects, they can recognize these objects even if they have seen them only once or a few times. Consequently, few-shot object detection (FSOD) naturally comes to an attractive research direction in this context \cite{jiaxu2021comparative,xiong2023robotic}. Specifically, FSOD aims to detect unfamiliar objects using only a few annotated examples, following pre-training in the initial phase on a vast amount of publicly accessible data, as illustrated in Figure. \ref{figure1} (a-b). This approach thereby reduces the burden of annotating a substantial volume of data in the target domain.

To achieve FSOD, current approaches focus on utilizing meta-learning \cite{liu2021afd,yin2020meta,chen2021ldanet}, transfer learning \cite{chen2018lstd,li2012review,yang2020context}, and data augmentation \cite{zhang2021hallucination,demirel2023meta} to fine-tune a base model and adapt the model to novel categories, while demonstrating some degree of success. Almost all existing FSOD methods require to access model parameters and leverage given object instances to fine-tune the model. Unfortunately, in some realistic scenarios, e.g., the ones encountered in industry applications \cite{huang2023industrial}, the constructed model necessitates swift/real-time configuration, or even a black-box model may be the only feasible option due to privacy preservation concerns \cite{theagarajan2021privacy}.  These limitations render model fine-tuning impractical for many current FSOD approaches. Consequently leading to inadequate model adaptation to new categories. Very naturally, a straightforward ambition is to investigate a novel task for FSOD, i.e., our proposed Test TIme Few Shot DEtection (TIDE)  without fine-tuning as shown in Figure. \ref{figure1} (c). 
It is worth emphasizing that the test time constraint adds an extra layer of complexity to the inherently challenging problem of FSOD. The reasons are as follows: (i) The determination of test time precludes the possibility of refining the model and thus limits its generalization towards novel classes. (ii) In conventional FSOD, the occurrence of data drift between the novel class samples and the base class samples can be mitigated through fine-tuning. Nonetheless, in the test-time setting, such an approach is unavailable.

To address these challenges,  we propose a novel FSOD approach, referred to as Test Time Few Shot DEtection (TIDE), which obviates the requirement for model fine-tuning during the configuration process. Specifically, considering the fact that the model fine-tuning is primarily attributed to the demand for new classification heads to accommodate novel categories, we introduce a practicable solution that provides a dynamic classifier associated with the support instance to avoid model fine-tuning. Moreover, to alleviate the potential mode collapse risk \cite{lala2018evaluation,yao2022discriminative} caused by the symmetric data encoder for support/query instances, we conduct an asymmetric architecture in our TIDE framework. Further, drawing inspiration from the promising advancements in Transformer-based detectors \cite{li2022dn,li2023mask,liu2017referring,wang2022framework,fellner2022data}, we present a robust feature detector built upon the Swin-Transformer for query instances \cite{zhang2022dino,liu2023grounding,liu2021swin} in this study. Besides, motivated by the intuition that good teachers can teach better students, we conduct an eminent support feature encoder, i.e., DINO \cite{caron2021emerging,oquab2023dinov2} provided by Facebook, to comprehensively investigate the semantic information conveyed by the support instances. In the end, a cross-attention module is incorporated to further enhance the performance of the dynamic classifier.

By formulating the above structures, we design a final TIDE framework for the test time FSOD. What makes our method most different from existing FSOD methods is that the dynamic classifier module plays an indispensable role in providing a more flexible way to adapt to novel categories.  In summary, our contributions are threefold:

\begin{itemize}
\item A novel FSOD scenario is presented, which offers a more realistic setting for test-time  FSOD (namely TIDE) without fine-tuning;
\item An effective FSOD method based on an asymmetric encoder is provided, which is the first attempt, to our best knowledge, to address TIDE without model fine-tuning;
\item An impressive experimental result reveals that the proposed TIDE methods can effectively enhance the FSOD ability of the model and are even better than the results of the fine-tuning methods.

\end{itemize}

The rest of this paper is organized as follows. In Section \ref{section2}, we briefly overview FSOD and Test Time Learning and highlight their difference compared to TIDE. In Section \ref{section3}, we elaborate on the problem formulation. In Section \ref{section4}, we present our TIDE model and its details. In Section \ref{section5}, we present the experimental results and the corresponding analysis. In the end, we conclude the entire paper with future research directions in Section \ref{section6}.

\section{Related Works}\label{section2}
In this section, we present the most related research on FSOD and test time learning. We also highlight the main differences between them and our method.
\subsection{Few Shot Objection Detection}
As a promising strategy, few-shot object detectors can effectively generalize and achieve satisfactory performance using only a limited number of labeled novel images, by leveraging abundant data for base classes. Two main branches leading in FSOD are Transfer Learning \cite{kim2020few} and Meta-Learning \cite{xie2022multi}. Transfer-learning methods aim to identify the most effective learning strategy for general object detectors on a limited set of new images. Wang et al. \cite{wang2020frustratingly} proposed fine-tuning solely the last layer using a classifier based on cosine similarity, while MPSR \cite{wu2020multi} addressed the issue of restricted scale diversity by incorporating a manually-defined positive refinement branch. A recent study \cite{zhu2021semantic} explored semantic relations between novel-base classes and introduced contrastive proposal encoding techniques. Meta-learning approaches typically consist of two branches: one for extracting support information and another for detecting objects in query images, with Meta R-CNN \cite{yan2019meta} and FSDet \cite{xiao2022few} focusing on support-guided query channel attention. To improve the small instance, A-RPN \cite{fan2020few} was provided by leveraging novel attention RPN and multi-relation classifiers.

Compared to the conventional FSOD methods, our derived TIDE approach does not need to fine-tune the model and thereby accelerating its configuration process. Consequently, TIDE is more applicable for \textbf{real-time model configuration}.
\subsection{Test Time Learning}

With the demand of rapid model configuration, test time learning is provided to avoid altering the additional training stage or finetuning all the layers in the test stage \cite{sun2020test,zhang2023domainadaptor}. In the Transfer learning area, several attempts focus on fast and online adaptation. e.g., test time training \cite{sun2020test,wang2022continual} or test time adaptation \cite{wang2020tent}. In the realm of re-ranking procedure,  Hudson et al. \cite{hudson2023if} designed a cross-attention knowledge distillation loss for test-time re-ranking. In the context of  Identity-Preserving Portrait Generation, Shi et al., \cite{shi2023instantbooth} learned rich visual feature representation by introducing a few adapter layers to the pre-trained mode for test-time fine-tuning.

The current Test time learning mainly focuses on model adaptation for pattern recognition. Besides, most of their work still \textbf{requires test-time fine-tuning}. In contrast, this is the first time, to our best knowledge, to address test time FSOD. Moreover, our derived TIDE  \textbf{does not require to access any model parameters} and thus is naturally suitable for black-box scenarios.  

\section{Test Time Few Shot Object Detection}\label{section3}
Test Time Few Shot Object Detection is a critical and challenging task that involves the ability to recognize and locate new objects in an image that were not seen during the training phase, with only a limited proportion of supporting samples provided at test time or deployment \textbf{without re-training or fine-tuning}. This task is essential for a wide range of applications from autonomous robotics to augmented reality, where a system must adapt to unseen objects in its environment and detect them adequately. The detailed definitions are given as follows:
\subsection{ Problem Difinition}\label{sec:part1}
In test-time few-shot object detection, given an image dataset consisting of $B$ base classes $\mathbf{C_B}$ and $N$ novel ones $\mathbf{C_N}$, where $\mathbf{C_B}\cap \mathbf{C_N} = \emptyset$ is satisfied, our objective is to train a model capable of localizing and classifying objects of novel classes in $\mathbf{C_N}$ by only providing $K$ image samples per novel class at test time and ample image samples in $\mathbf{C_B}$ during training stage.
\subsection{Task}
Incorporating a wealth of base data from base classes during the training of the base model, the objective of test-time few-shot object detection is to achieve strong generalization capabilities on novel objects in a query image set, utilizing a few or even a single novel image shot, without any model fine-tuning.

\section{Method}\label{section4}
In this section, we first briefly summarize the object detection pipeline that we use as the basis for building our model. Then, we describe how we tackled this problem.
\subsection{TIDE}\label{sec:part2}
\begin{figure*}[ht]
\centering
\includegraphics[width=0.9\textwidth]{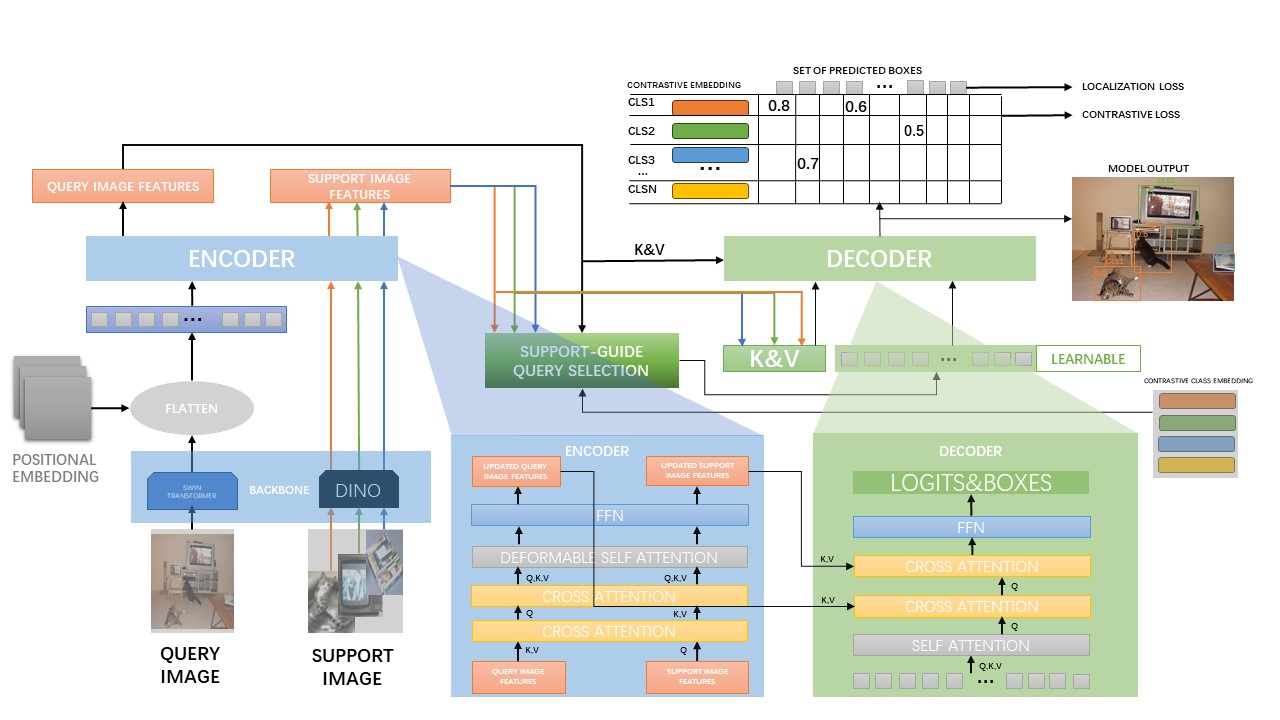}
\caption{TIDE exploits an asymmetric backbone which comprises Swin Transformer \cite{liu2021swin} for query image feature extraction and self-supervised vision transformer DINO \cite{caron2021emerging} for support image feature extraction. An encoder and a decoder comprising cross attention are applied to better relate query image features and support image features. Ultimately a dynamic contrastive classifier and a bounding box linear regressor are used for final object localization and classification.}
\label{fig_1}
\end{figure*}

Our proposed Test Time Few-Shot Object Detection (TIDE) is built upon the Transformer \cite{vaswani2017attention}. The entire pipeline of our derived TIDE is given in Figure. \ref{fig_1}.

The architecture of TIDE consists of 1) an asymmetric backbone with self-supervised vision transformer DINO\cite{caron2021emerging} and Swin Transformer\cite{liu2021swin}, 2) a dynamic contrastive classifier, 3) a transformer decoder that processes object queries selected by leveraging visual representations from the support image set. In a nominally larger departure from previous related works\cite{fan2020fsod,han2022few,li2022airdet}, Our model takes in support images belonging to novel classes together with a query image in a single forward at test time w/o any fine-tuning operation or retraining, and outputs both the location and classification on novel objects.

Specifically, given two different image sets, i.e., support set and query set, we retrieve their abstract visual representation with two different transformer vision encoders in the first place while query image features are downsampled, flattened, and attached with positional embeddings. To establish mutual visual communication between the query set and the support set, dual visual features are aligned with cross attention mechanism \cite{chen2023inventory,wu2023memory}. In the sequel, we use a support-guide query selection module to initialize object queries before the decoder with the assistance of aligned features. Similar to how learnable object queries are used in most DETR-like models, these object queries are fed into the decoder for final bounding box regression and classification.

\subsubsection{Feature Representation Extraction and Amplification}\label{sec:part2_1}
To avoid the potential mode collapse caused by the same feature encoder of support/query images, we design an asymmetric feature encoder.  Inspired by a natural intuition that effective teachers can facilitate better learning outcomes for students, we employ a SOTA feature encoder i.e., DINO \cite{caron2021emerging,oquab2023dinov2} for support encoder. Further, we present a robust feature detector built upon the Swin-Transformer for query instances \cite{zhang2022dino,liu2023grounding,liu2021swin} mainly due to its promising advancements \cite{li2022dn,li2023mask,liu2017referring,wang2022framework,fellner2022data}.

Specifically, considering two image (support, query) sets, we extract hierarchical query image features $\mathbf{Z}\in \mathbf{R}^{H \times W\times d}$ from query image set $\mathbf{Q_i} \in \mathbf{R}^{H^{\prime} \times W^{\prime} \times 3},i\in\{1, \ldots, m\}$ where $m$ denotes the number of classes(i.e., the number of support images) in current training iteration, $H, W, d$ represent height, width and hidden dimension respectively, utilizing Swin Transformer\cite{liu2021swin} with its outputs from different blocks while harnessing DINO\cite{caron2021emerging} for obtaining more explicit visual representation $\mathbf{S}\in \mathbf{R}^{m\times d}$ from support image set. Subsequently, query and support features (attached with fixed positional embedding following \cite{carion2020end}) are enhanced with cross attention by Bi-MultiHeadAttention layers (BMHA), Layer Normalization (LN), and Feed Forward Network (FFN) layers, which to a great extent ameliorates the performance of the proposed dynamic contrastive classifier.
\begin{equation}
\label{eq1}
\mathbf{Z}^\prime,\mathbf{S}^\prime=\mathbf{BMHA}(LN(\mathbf{Z}),LN(\mathbf{S}))
\end{equation}
Through the aid of BMHA, features in dual branches after being extracted from asymmetric backbones can seamlessly interact and fuse with information from both branches without affecting the original computational flow, shortening the similarity gap between target objects in the query images and the support images.
To capture long-range dependencies and enrich them with contextual information. We also take advantage of Deformable Multiscale self-attention(DMSA)\cite{zhu2020deformable,chen2020survey} to further form spatial relationships between image features of the two branches, as depicted in Figure.\ref{fig_1}:
\begin{equation}
\label{eq2}
\mathbf{Z}^{\prime\prime},\mathbf{S}^{\prime\prime}=\mathbf{DMSA}(LN(\mathbf{Z}^\prime),LN(\mathbf{S}^\prime))
\end{equation}
\begin{equation}
\label{eq3}
\mathbf{Z}^\mathbf{L},\mathbf{S}^\mathbf{L}=\mathbf{FFN}(LN(\mathbf{Z}^{\prime\prime}),LN(\mathbf{S}^{\prime\prime}))
\end{equation}
The TIDE's encoder, as depicted in Figure.\ref{fig_1}, is stacked with $\mathbf{L}$ layers that follow the formulations outlined in Eqs.\ref{eq1},\ref{eq2} and \ref{eq3}. Note that $\mathbf{L}$ is set to 6 in our implementation.
Since TIDE aims to locate and classify specific objects referenced in the support set, we also guide the initialization of object queries $\mathbf{O}\in \mathbf{R}^{\mathcal{N}\times \mathbf{d}}$, by selecting features from $\mathbf{Z}^{L}$ that are more pertinent to the ones extracted from support images, the selected features are represented by acquiring the indexes $\mathbf{I}\in \mathbf{R^\mathcal{N}}$. Notably, TIDE infers a fixed set of $\mathcal{N}$ candidate predictions in a single pass through the decoder. While $\mathcal{N}$  is significantly larger than the typical number of objects in an image,  the remaining unselected object queries $\mathbf{O}^\prime$ are set to be learnable by the model during the training phase. The detailed processes are given below:
\begin{equation}
\label{eq4}
\mathbf{Z}^{L^\prime}= \mathbf{Z}^{L}\times \mathbf{S}^\top
\end{equation} 
\begin{equation}
\label{eq5}
\mathbf{Z}^{L^{\prime\prime}}=\max \mathbf{Z}^{L^\prime}
\end{equation}
\begin{equation}
\label{eq6}
\mathbf{I}=\mathbf{\tau}\left(\mathbf{Z}^{L^{\prime\prime}},k,\mathbf{\zeta}\right)
\end{equation}
\begin{equation}
\label{eq7}
\mathbf{O}=\left[\begin{array}{ll}
\mathbf{O_I} & \mathbf{O}^{\prime}
\end{array}\right]
\end{equation}
where $\mathbf{\tau}$ is a function that returns the $k$ largest elements of the given input matrix along a given dimension $\zeta$.

\subsubsection{TIDE decoder}\label{sec:part2_2}
The decoder of TIDE takes in the selected and learnable object queries $\mathbf{O}$ while simultaneously adopting both query and support image features $\mathbf{Z}^\mathbf{L},\mathbf{S}^\mathbf{L}$ from encoder output as keys and values for extra dual cross attention after Multi-Head Self-Attention(MHSA) as depicted in Figure.\ref{fig_1} in order to impart positive object information from support images to query image's representation:
\begin{equation}
\mathbf{C}^{\prime}  =\operatorname{MHSA}\left(\operatorname{LN}\left(\mathbf{C}^{L-1}\right)\right)+\mathbf{C}^{L-1}
\end{equation}
\begin{equation}
\mathbf{C}^{\prime \prime}  =\operatorname{MHCA}\left(\operatorname{LN}\left(\mathbf{C}^{\prime}\right),\operatorname{LN}\left(\mathbf{Z}^L\right)\right))+\mathbf{C}^{\prime}, 
\end{equation}
\begin{equation}
\mathbf{C}^{\prime \prime\prime}  =\operatorname{MHCA}\left(\operatorname{LN}\left(\mathbf{C}^{\prime\prime}\right), \operatorname{LN}\left(\mathbf{S}^L\right)\right)+\mathbf{C}^{\prime\prime}, 
\end{equation}
\begin{equation}
\mathbf{C}^L
=\operatorname{FFN}\left(\operatorname{LN}\left(\mathbf{C}^{\prime \prime\prime}\right)\right)+\mathbf{C}^{\prime \prime\prime},
\end{equation}
\begin{figure}[ht]
\centering
\includegraphics[width=0.48\textwidth]{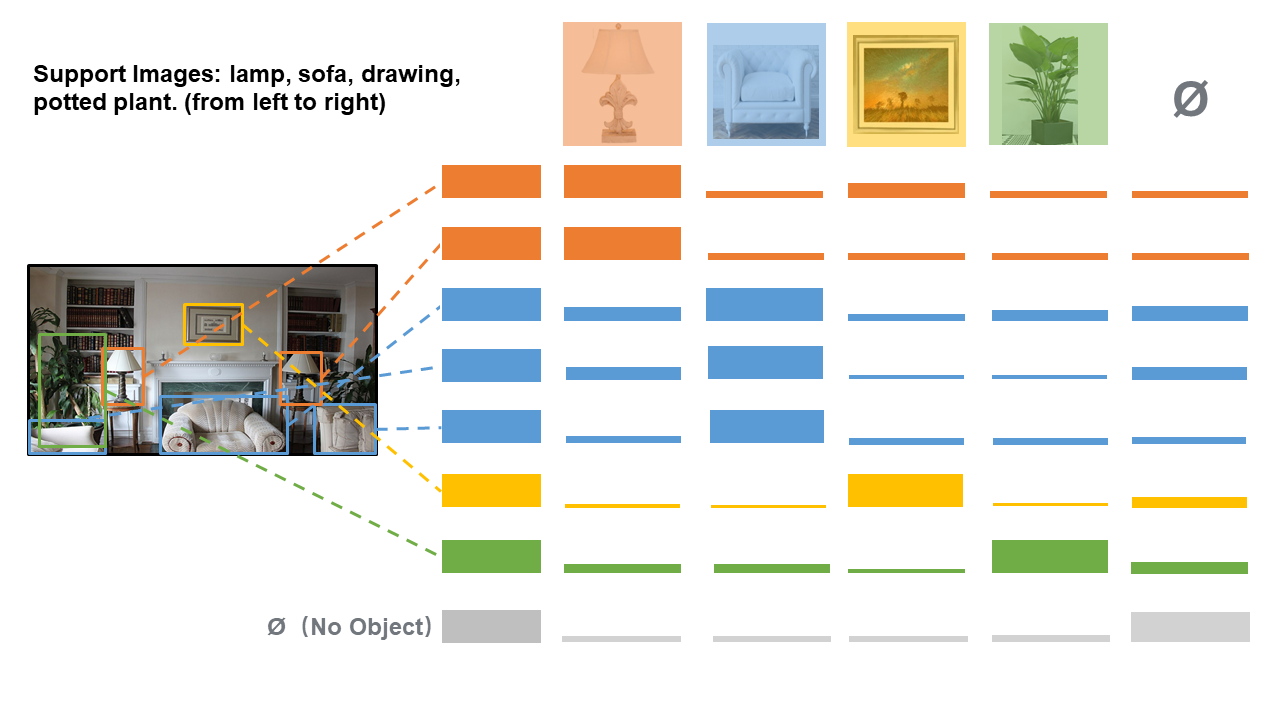}
\caption{Illustration of the dynamic contrastive classifier. For each object, the model predicts a distribution over the support images' positions in the input sequence.}
\label{figure3}
\end{figure}where $\mathbf{C}^{L=0} = \mathbf{O}, \mathbf{O}\in R^{\mathcal{N}\times d}$ and $C^\prime,C^{\prime\prime},C^{\prime\prime\prime}$ represent temporary variables for calculating $C^L$ while specifically in TIDE's implementation, the decoder is stacked with $L=6$ layers. Further, a Multi-Head Cross Attention (MHCA) Module from Deformanble-DETR \cite{zhu2020deformable} is also incorporated to further enhance the final performance of the dynamic classifier. 
\subsubsection{Dynamic Contrastive Classifier}\label{sec:part2_3}
Conventional DETR-like object detectors \cite{li2022dn,liu2022dabdetr,zhu2020deformable} rely on a predetermined class embedding to represent and distinguish among various object classes, which significantly limits their application to closed-set detection solely. In contrast, we propose a dynamic contrastive classifier to enhance its compatibility with test-time few-shot object detection. In particular, the classifier predicts the distribution $\hat{\mathbf{P}}$ determined by considering the positional relationship of each support image corresponding to the target objects as depicted in Figure.\ref{figure3}. The detailed formulation is given as follows:
\begin{equation}
\hat{\mathbf{P}} = \operatorname{Softmax}\left(\mathbf{C}^L\mathbf{S}^\top\right)
\end{equation}
The derived $\hat{\mathbf{P}}$ to a great extent guaranteed the accuracy of TIDE in downstream bounding box detection tasks by introducing a similarity computation between two continuous numerical representations. As a result, it's capable of generalizing on novel objects based on their similarity with the support image set.
\begin{figure}[ht]
\centering
\includegraphics[width=0.48\textwidth]{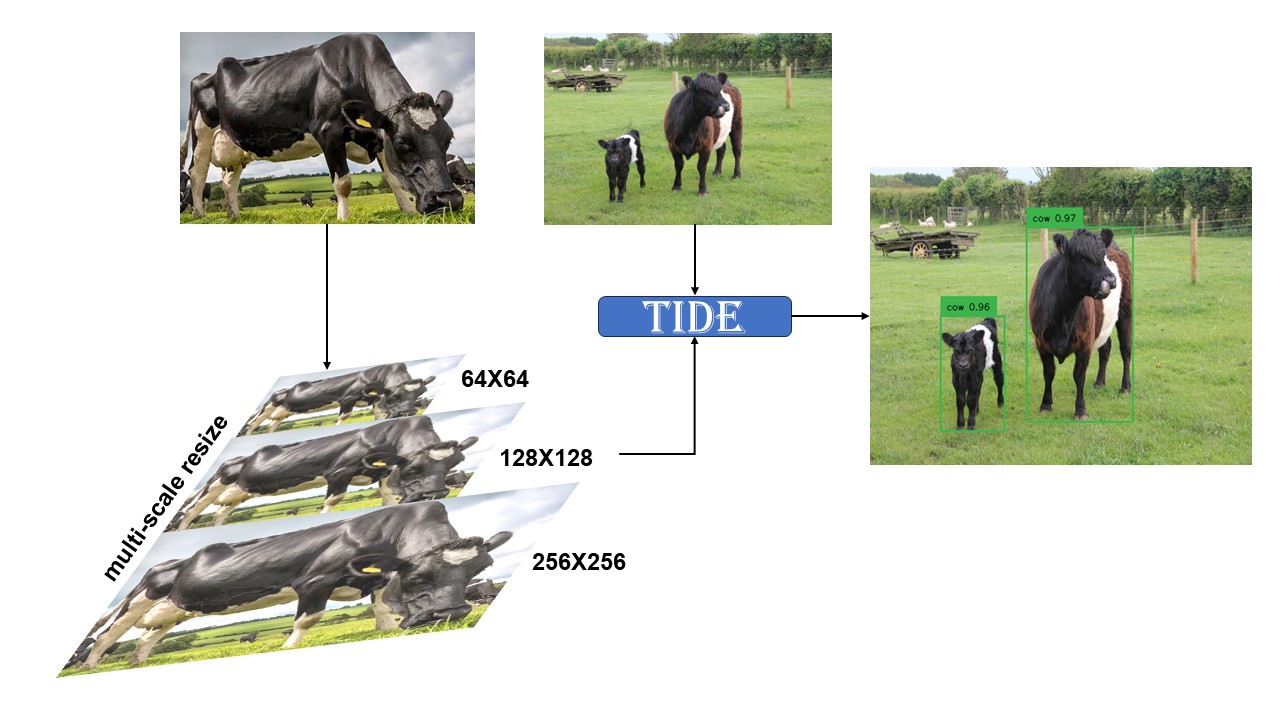}
\caption{Depiction of multi-scale resizer for support images. Resizing support images into multiple predetermined shapes enables TIDE to capture more coarse-to-fine-grained details which predominantly improves TIDE's performance.   }
\label{figure:multi-scale}
\end{figure}

\subsection{Training and Loss Functions}\label{sec:part2_4} 
During the training phase, for each base class that exists in the query image, first, we generate a support image set by randomly sampling and cropping objects with bounding box annotations from that base class employing a different image from the training dataset while filling in an arbitrary proportion of negative samples by cropping random objects from different classes to fortify the model's robustness to distinguish among different objects.

Given the specificity of our dynamic contrastive classifier, we then make sure in each training iteration each base class of the objects in the query image solely corresponds to one single support image, e.g., suppose a query image that contains three dogs and five cats, we only provide two support images respectively indicating a dog and a cat during training phase since the model's final class prediction accords to relative positions of each support image. Nevertheless, at test time our model naturally supports one-way multiple shots. For example, when provided with two support images of the same class, the model exerts the main effort to draw a parallel between target objects and support images then classifies and locates objects that are most similar to the specific single support image while both support images defacto manifest the same class. Thus, for each query image and support image $\mathbf{i}$ of positive samples, the ground truth $\mathbf{y}_i = (\mathbf{\xi}_i,\mathbf{\delta}_i)$, where $\mathbf{\xi}_i$ indicates the target label determined by the current support image's relative position and $\mathbf{\delta}_i\in [0,1]^4$ denotes the normalized target bounding box coordinates. To calculate final loss functions for our model, the contrastive classification head and bounding box regression head are exploited to predict a set of $\mathcal{N}$ predictions $\hat{\mathbf{y}}=\left\{\hat{\mathbf{y}}_i\right\}_{i=1}^\mathcal{N}$ comprising class probabilities $\hat{\mathbf{p}}_i\left(\mathbf{\xi}^s\right)$ and bounding box coordinates $\hat{\mathbf{\delta}}_i$. Concretely, the contrastive classification head is implemented with 6 layers of contrastive embedding and the bounding box regression head consists of a 6-layer MLP and ReLU activations. 

Akin to the original DETR but with slight changes \cite{carion2020end}, we use a special no object class $\varnothing$ to denote invalid predictions and bipartite matching to find an optimal permutation $\left\{\hat{\mathbf{y}}_{\mathbf{\sigma} i}\right\}_{i=1}^\mathcal{N}$. Thus, the final loss can be defined as follows:
\begin{equation}\begin{aligned}
\mathbf{L}= & \sum_{i=1}^\mathcal{N} \lambda_1 \mathbf{L}_{\mathrm{CrossEntropy}}\left(\mathbf{\xi}_i^s, \hat{\mathbf{p}}_{\mathbf{\sigma}(i)}\left(\mathbf{\xi}^s\right)\right) \\
& +\lambda_2\left\|\mathbf{\delta}_i-\hat{\mathbf{\delta}}_{\mathbf{\sigma}(i)}\right\|_1+\lambda_3 \operatorname{IoU}\left(\mathbf{\delta}_i, \hat{\mathbf{\delta}}_{\mathbf{\sigma}(i)}\right),
\end{aligned}\end{equation}
where IoU is the GIoU loss \cite{rezatofighi2019generalized} and $\lambda_i$ are the corresponding loss weights for balancing different regularizer. Following DETR {\cite{carion2020end}, we also empirically set $\lambda_1=1$, $\lambda_2=5$ and $\lambda_3=2$.
\subsection{Multi-scale Resizer for Support Image}
To mitigate the impact of missed detections due to image size variations. At test time we also extract multi-scale(MS) features from support images by resizing operation on support images to [64$\times$64, 128$\times$128, 256$\times$256] shown in Figure.\ref{figure:multi-scale} and a fix rescaling to 128$\times$128 for batch alignment.
\begin{table}[]
\tiny
\centering
\caption{Statistics of the benchmark datasets used in training and evaluating TIDE}
\begin{tabular}{ccccc}
\hline\hline
Dataset        & \#Base Sample & \#Base Class & \#Novel Sample & \#Novel Class \\ \hline\hline
COCO           & 91888         & 60           &  5000          &  20   \\ 
PASCAL VOC     & -          & -           &  5717          &  20   \\ \hline\hline
\end{tabular}

\label{table1}
\end{table}

\section{Experiment}\label{section5}
To showcase the performance of TIDE, a series of experiments are conducted. Section \ref{section5_1} describes the setup, followed by the experimental results in Section \ref{section5_2}.
\subsection{Setup}\label{section5_1}
The \emph{COCO} 2017 dataset is adopted to train the base model of TIDE on an NVIDIA A800 GPU, where small-size images are filtered out. The dataset is also the source of support images. In particular, the images that correspond to the categories of objects in query images (i.e., $\mathbf{C_N}$ defined in Section \ref{section3}) are first cropped from the bounding box annotation area and then sorted based on their respective categories. Thus, the \emph{COCO} dataset is divided into two subsets: the novel categories, $\mathbf{C_N}$, and the base ones that are part of \emph{COCO} classes but not in $\mathbf{C_N}$.  For each training iteration, the support set consists of a number of positive samples that are randomly extracted from the cropped images. In addition, an extra no-object image is considered as one of the negative samples that can be created using $numpy.zeros((128,128,3))$. All images in the support set are then resized to 128 $\times$ 128 pixels.  To increase the difficulty of object detection, augmentation methods, such as random color jitter and random horizontal flip, are applied to the support images.

 In addition to \emph{COCO}, \emph{PASCAL VOC} dataset is used to evaluate the performance of TIDE. The aforementioned process of establishing support images is then applied to $\mathbf{C_N}$. As shown in Table \ref{table1}, the two datasets share the same 20 novel categories.

\subsection{Evaluation setting}\label{section5_2}
In the first set of experiments, we adopt the 2-way N-shot $({N = 1,2,3,4,5})$ evaluation method \cite{li2022airdet,fan2020fsod}, and extensively test and compare TIDE to the state-of-the-art FSOD approaches, including the ones without fine-tuning(i.e., A-RPN\cite{fan2020few} and AirDet\cite{li2022airdet}),  and with fine-tuning (i.e., FRCN\cite{ren2015faster}, TFA$_{fc}$\cite{wang2020frustratingly}, TFA$_{cos}$\cite{wang2020frustratingly}, FSDetView\cite{xiao2022few}, MPSR\cite{wu2020multi}, W. Zhang, et al.\cite{zhang2021hallucination} and FADI\cite{chen2018lstd}). In addition, we look at the performance difference between TIDE with and without (w/wo) multi-scale (MS). Note that all these methods are tested on both the cross-domain dataset \emph{PASCAL VOC} 2012 and in-domain dataset \emph{COCO} 2017 under the same evaluation setting  \cite{wang2020frustratingly,li2022airdet,xiao2022few}, where data is split following the strategies presented in \cite{kang2019few,wang2020frustratingly}, and different seeds of the split data are used for multiple runs.


\begin{table*}[]
\centering
\caption{Few-shot object detection results on the COCO dataset.}
\begin{tabular}{cc|ccc|ccc|ccc|ccc}
\hline\hline
Shots                &   &          & \multicolumn{1}{c}{1} &           &      & \multicolumn{1}{c}{2} &           &      & \multicolumn{1}{c}{3} &           &      & 5         &           \\ 
Method               & \multicolumn{1}{l|}{Finetune}  & $AP$     & $AP_{50}$             & $AP_{75}$ & $AP$ & $AP_{50}$             & $AP_{75}$ & $AP$ & $AP_{50}$             & $AP_{75}$ & $AP$ & $AP_{50}$ & $AP_{75}$ \\ \hline\hline
A-RPN\cite{fan2020few}                & \XSolidBrush    &$4.32$ &$7.62$ &$4.30$ &$4.67$ &$8.83$ &$4.49$ &$5.28$ &$9.95$ &$5.05$ &$6.08$ &$11.17$&$5.88$  \\
AirDet\cite{li2022airdet}               & \XSolidBrush    &$5.97$ &$10.52$&$5.98$ &$6.58$ &$12.02$&$6.33$ &$7.00$ &$12.95$&$6.71$ &$7.76$ &$14.28$&$7.31$  \\
TIDE(w/o MS)         & \XSolidBrush    &$5.93$ &$10.33$&$5.95$ &$8.24$ &$13.52$&$8.56$ &$9.36$ &$15.28$&$9.53$ &$10.77$&$18.06$&$11.24$  \\ 
TIDE                 & \XSolidBrush    &$\textbf{9.87}$ &$\textbf{16.16}$&$\textbf{10.51}$&$\textbf{12.55}$&$\textbf{21.29}$&$\textbf{13.26}$&$\textbf{12.35}$&$\textbf{20.81}$&$\textbf{12.73}$&$\textbf{11.75}$&$\textbf{19.89}$&$\textbf{12.01}$  \\ \hline
FRCN\cite{ren2015faster}                 & \CheckmarkBold  &$3.26$ &$6.66$ &$3.04$ &$3.73$ &$7.79 $&$3.22$ &$4.59$ &$9.52 $&$4.07$ &$5.32$ &$11.20$&$4.54$\\
TFA$_{fc}$\cite{wang2020frustratingly}           & \CheckmarkBold  &$2.78$ &$5.39$ &$2.36$ &$4.14$ &$7.98 $&$4.01$ &$6.33$ &$12.10$&$5.94$ &$7.92$ &$15.58$ &$7.29$\\
TFA$_{cos}$\cite{wang2020frustratingly}          & \CheckmarkBold  &$3.09$ &$5.24$ &$3.21$ &$4.21$ &$7.70 $&$4.35$ &$6.05$ &$11.48$&$5.93$ &$7.61$ &$14.56$&$7.17$\\
FSDetView\cite{xiao2022few}            & \CheckmarkBold  &$2.20$ &$6.20$ &$0.90$ &$3.40$ &$10.00$&$1.50$ &$5.20$ &$14.70$&$2.10$ &$8.20$ &$21.60$&$4.70$\\
MPSR\cite{wu2020multi}                 & \CheckmarkBold  &$3.34$ &$6.11$ &$3.25$ &$5.41$ &$9.68 $&$5.52$ &$5.70$ &$10.54$&$5.50$ &$7.20$ &$13.55$&$6.89$\\
A-RPN\cite{fan2020few}                & \CheckmarkBold  &$4.59$ &$8.85$ &$4.37$ &$6.15$ &$12.05$&$5.76$ &$8.24$ &$15.52$&$7.92$ &$9.02$ &$17.29$&$8.53$\\
W.Zhang et al.\cite{zhang2021hallucination}       & \CheckmarkBold  &$4.40$ &$7.50$ &$4.90$ &$5.60$ &$9.90 $&$5.90$ &$7.20$ &$13.30$&$7.40$ &$-$    &$-$    &$-$\\
FADI\cite{chen2018lstd}                 & \CheckmarkBold  &$5.70$ &$10.40$&$6.00$ &$7.00$ &$13.01$&$7.00$ &$8.60$ &$15.80$&$8.30$ &$10.10$&$18.60$&$11.90$\\
AirDet\cite{li2022airdet}               & \CheckmarkBold  &$6.10$ &$11.40$&$6.04$ &$8.73$ &$16.24$&$8.35$ &$9.95$ &$19.39$&$9.09$ &$10.81$&$20.75$&$10.27$\\ 
\hline\hline
\label{tableCOCO}

\end{tabular}
\end{table*}

\begin{table*}[]
\centering
\caption{Few-shot object detection results on the PASCAL VOC dataset.}
\begin{tabular}{cc|ccc|ccc|ccc|ccc}
\hline\hline
Shots                &   &          & \multicolumn{1}{c}{1} &           &      & \multicolumn{1}{c}{2} &           &      & \multicolumn{1}{c}{3} &           &      & 5         &           \\ 
Method               & Finetune        & $AP$         & $AP_{50}$    & $AP_{75}$     & $AP$        & $AP_{50}$    & $AP_{75}$    & $AP$         & $AP_{50}$     & $AP_{75}$     & $AP$         & $AP_{50}$    & $AP_{75}$ \\ \hline\hline
A-RPN\cite{fan2020few}                & \XSolidBrush    &$10.45$&$18.10$&$10.32$&$13.10$&$22.60$&$13.17$&$14.05$&$24.08$&$5.05$ &$15.87$&$35.03$&$15.26$ \\
AirDet\cite{li2022airdet}               & \XSolidBrush    &$11.92$&$21.33$&$11.56$&$15.80$&$26.80$&$16.08$&$16.08$&$28.61$&$6.71$ &$17.83$&$29.78$&$18.38$  \\
TIDE(w/o MS)         & \XSolidBrush    &$12.56$       &$24.16$       &$11.72$       &$14.88$       &$27.20$       &$14.58$       &$15.90$       &$26.34$       &$16.37$       &$16.12$       &$28.48$       &$15.79$       \\ 
TIDE                 & \XSolidBrush    &$\textbf{16.88}$&$\textbf{26.81}$&$\textbf{17.49}$&$\textbf{19.67}$&$\textbf{33.02}$&$\textbf{20.17}$&$\textbf{20.90}$&$\textbf{35.15}$&$\textbf{21.36}$&$\textbf{21.36}$&$\textbf{37.06}$&$\textbf{21.31}$  \\ \hline
FRCN\cite{ren2015faster}                 & \CheckmarkBold  &$4.49$ &$9.44$ &$3.85$ &$5.20$ &$11.92$&$3.84$ &$6.50$ &$14.39$&$5.11$ &$6.55$ &$14.48$&$5.09$\\
TFA$_{cos}$\cite{wang2020frustratingly}          & \CheckmarkBold  &$4.66$ &$7.97$ &$5.14$ &$6.59$ &$11.91$&$6.49$ &$8.78$ &$17.09$&$8.15$ &$10.46$&$20.93$&$9.53$\\
TFA$_{fc}$\cite{wang2020frustratingly}           & \CheckmarkBold  &$4.40$ &$8.60$ &$4.21$ &$7.02$ &$13.80$&$6.21$ &$9.24$ &$18.48$&$8.03$ &$11.11$&$22.83$&$9.78$\\
FSDetView\cite{xiao2022few}            & \CheckmarkBold  &$4.80$ &$14.10$&$1.40$ &$3.70$ &$11.60$&$0.60$ &$6.60$ &$22.00$&$1.20$ &$10.80$&$26.50$&$5.50$\\
MPSR\cite{wu2020multi}                 & \CheckmarkBold  &$6.01$ &$11.23$&$5.74$ &$8.20$ &$15.08$&$8.22$ &$10.08$&$18.29$&$9.99$ &$11.49$&$21.33$&$12.06$\\
A-RPN\cite{fan2020few}                & \CheckmarkBold  &$9.49$ &$17.41$&$9.42$ &$12.71$&$23.66$&$12.44$&$14.89$&$26.30$&$14.76$&$15.09$&$28.08$&$14.17$\\
AirDet\cite{li2022airdet}               & \CheckmarkBold  &$13.33$&$24.64$&$12.68$&$17.51$&$30.35$&$17.61$&$17.68$&$32.05$&$17.34$&$18.27$&$33.02$&$17.69$ \\ 
\hline\hline
\label{tablepascal}

\end{tabular}
\end{table*}
\subsection{Experimental Results}

The comparison results on \emph{COCO} and \emph{PASCAL VOC} are then shown in Tables \ref{tableCOCO} and \ref{tablepascal}, respectively. The Boldface in these tables denotes the best performance for each line. From
these results, we can easily observe that the proposed TIDE significantly outperforms the existing FSOD methods without finetuning. Specifically, on the one-shot task, TIDE, in comparison to AirDet,  achieves 65.3\%, 53.6\% and 75.8\% on improvement in Average Precision (AP), AP50, and AP75 on the \emph{COCO} dataset. Similar results appear on the PASCAL VOC dataset, showing the enhancement of  41.6\%, 25.6\% and 33.9\%. More importantly, our method is even better than the ones with finetuning across all tasks, which attests to the superiority of the proposed TIDE. The comparison between TIDE w/wo MS further corroborates the efficacy of the multi-scale technique for supporting images.

The second set of experiments evaluates the performance of the derived cross-attention module. As clearly seen in the feature maps of the cross-attention layer in Figure \ref{figure:atten}, attention is paid to the objective area, which demonstrates the fact that our cross-attention module facilitates the model to learn more comprehensive contextual information and distinctive features, thereby enhancing its performance.
\begin{figure}[ht]
\centering
\includegraphics[width=0.5\textwidth]{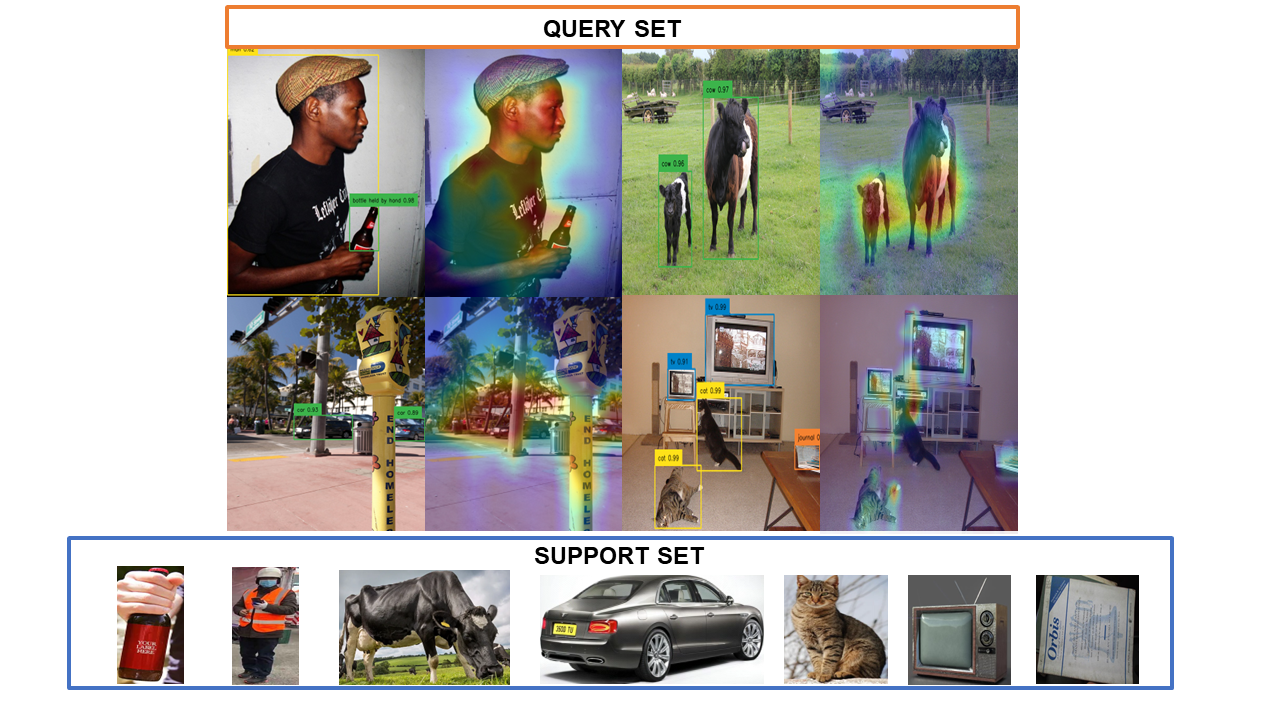}
\caption{Heat map visualization of cross attention module. In virtue of support-query cross-attention at the feature-enhancing stage, TIDE is capable of focusing on the target objects and noticing their representative parts which eventually induces accurate classification and localization in downstream detection tasks.}
\label{figure:atten}
\end{figure}
The proposed TIDE is composed of two primary components (i.e., BMHA and DCC). To investigate the sensitivity of the derived components on TIDE’s performance, the ablation experiments are conducted on the \emph{COCO} dataset. Specifically, we run TIDE, TIDE w/o BMHA, and TIDE w/o DCC \footnote{After DCC is removed, a binary class embedding of target and non-target is used as the classifier.} on \emph{COCO} with different novel sets. As we observe in Table.\ref{tab:ablation}, the performance dramatically declines when any of the BMHA and DCC components are removed. The results reveal the indispensability of the proposed two components.
\begin{table}[]
    \centering
    \caption{Abltation Study on BMHA and DCC for evaluating TIDE}
    \begin{tabular}{c|cccc}
\hline\hline
Approach  & \multicolumn{4}{c}{ COCO Novel Set } \\
& 1 & 2 & 3 & 5 \\ \hline\hline
 TIDE &\textbf{9.87}  &\textbf{12.55}  &\textbf{12.35}  &\textbf{11.75}  \\
TIDE w/o BMHA  &3.55  &4.62  &5.60  &6.96  \\
TIDE w/o DCC &5.12  &7.24  &8.31  &9.25  \\
\hline\hline
\end{tabular}
        \label{tab:ablation}
\end{table}

\section{Conclusion}\label{section6}
In responding to the demands of Industry 5.0 applications for real-time model configuration and privacy protection (black box), we focus on the test time FSOD, i.e., TIDE. Specifically, to mitigate the potential risk of mode collapse, we develop an asymmetric architecture, where the support branch and query branch possess different feature extractors. Furthermore, inspired by a natural intuition that a proficient teacher inherently fosters capable students, we employ a state-of-the-art (SOTA) feature extractor, i.e., DINO, to thoroughly explore the semantic feature of the support images. To circumvent model fine-tuning, we design a dynamic classifier associated with the support instance. Moreover, a cross-attention module is implemented to bolster the efficacy of the dynamic classifier. To avoid the impact of the target scale of the instances, we introduce a multi-scale resizer to further improve the model performance. Experimental results validate the effectiveness of the proposed TIDE approach. It is worth noting that while we emphasize the importance of semantic extraction in object detection, this aspect is not considered in this work. We would like to devise a general coarse-to-fine-grained semantic feature optimization strategy in the future, tailored closely to the unique constraints of Industry 5.0 applications, which enhances model performance and promotes superior generalization effects. 

\section*{Acknowledgments}
The authors would like to thank Dr. Wenming Cao, Dr. Huayang Tang, Dr. Miaolong Ye, Dr. Meng Tan, and Dr. Lei Li for the proofreading of this manuscript and for their technical support.

This work was partly supported by the  National Natural Science Foundation of China (62306051), the Fundamental Research Funds for the Central Universities (NJ2022028) and the Scientific and the Technological Research Program of Chongqing Municipal Education Commission (KJQN202300718).


\bibliographystyle{ieeetr}
\bibliography{mybibfile}

\newpage

\vspace{11pt}

\begin{IEEEbiography}[{\includegraphics[width=1in,height=1.25in,clip,keepaspectratio]{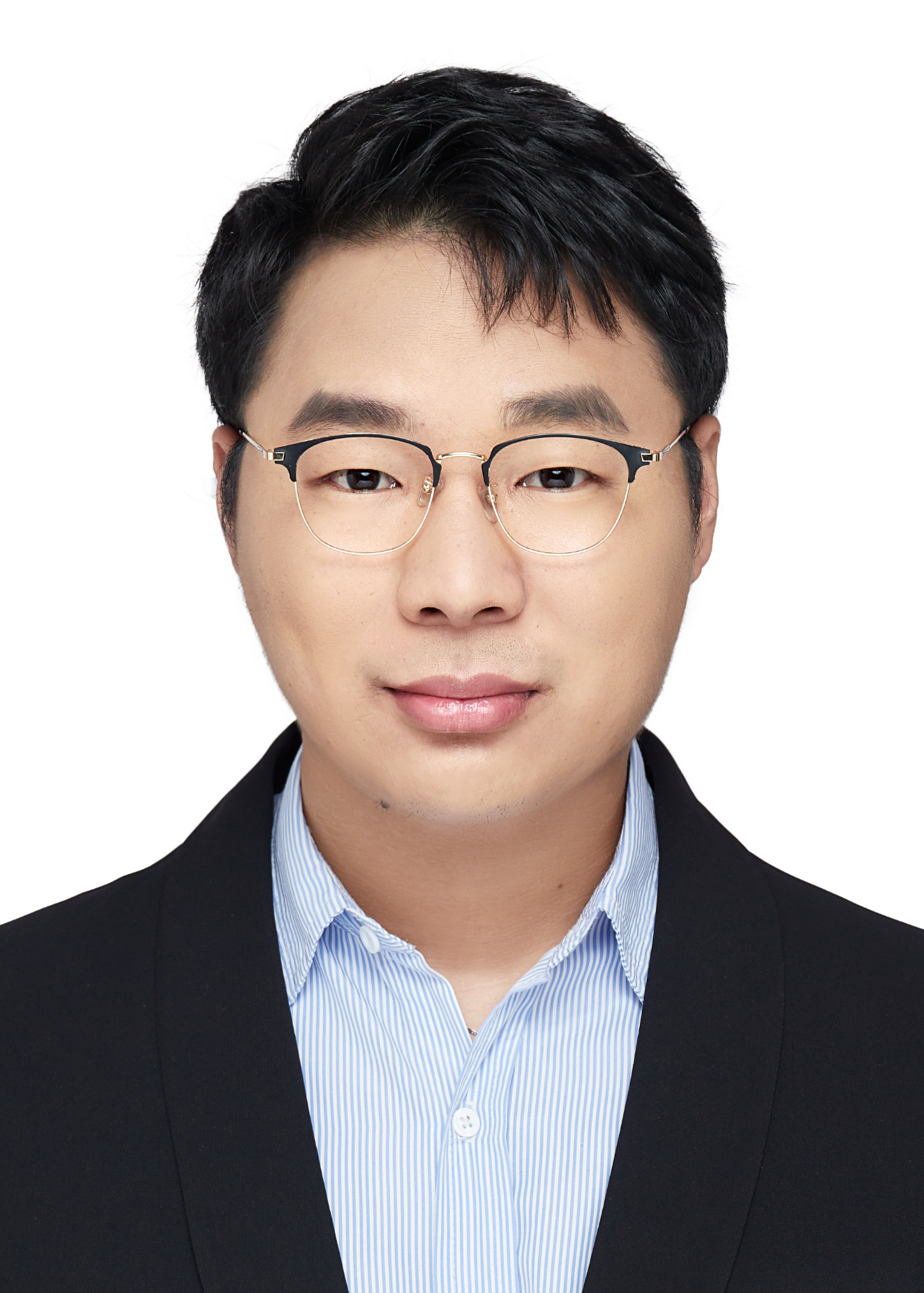}}]{Weikai Li}
received his B.S. degree in Information and Computing Science from Chongqing Jiaotong University in 2015. In 2018, he completed his M.S. degree in computer science and technique at Chongqing Jiaotong University.  In 2022, he completed his Ph.D. degree with the College of Computer Science \& Technology, Nanjing University of Aeronautics and Astronautics. He is now a lecturer at the School of Mathematics and Statistics, Chongqing Jiaotong University. His research interests include pattern recognition and machine learning.
\end{IEEEbiography}
\begin{IEEEbiography}[{\includegraphics[width=1in,height=1.25in,clip,keepaspectratio]{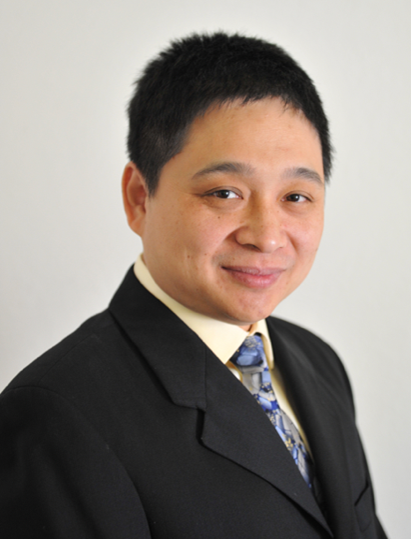}}]{Hongfeng (Jason) Wei } is currently visiting professor at the University of California, Los Angeles and Irvine; He is the CEO of China Science IntelliCloud Technology. He is also the Senior consultant of Pfizer Pharmaceuticals, Amazon and General Electric; Served as CTO of Nasdag listed technology company.
\end{IEEEbiography}
\begin{IEEEbiography}
[{\includegraphics[width=1in,height=1.25in,clip,keepaspectratio]{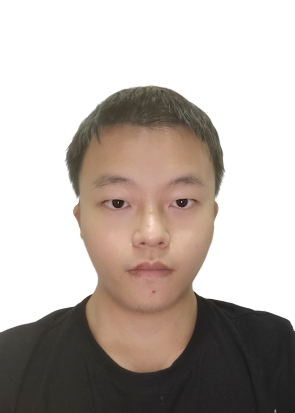}}]
{Yanlai Wu} is currently pursuing a B.S degree in Artificial Intelligence from the School of Information Science and Engineering, Chongqing Jiaotong University. His current research involves pattern recognition, computer vision and natural language processing.
\end{IEEEbiography}
\begin{IEEEbiography}[{\includegraphics[width=1in,height=1.25in,clip,keepaspectratio]{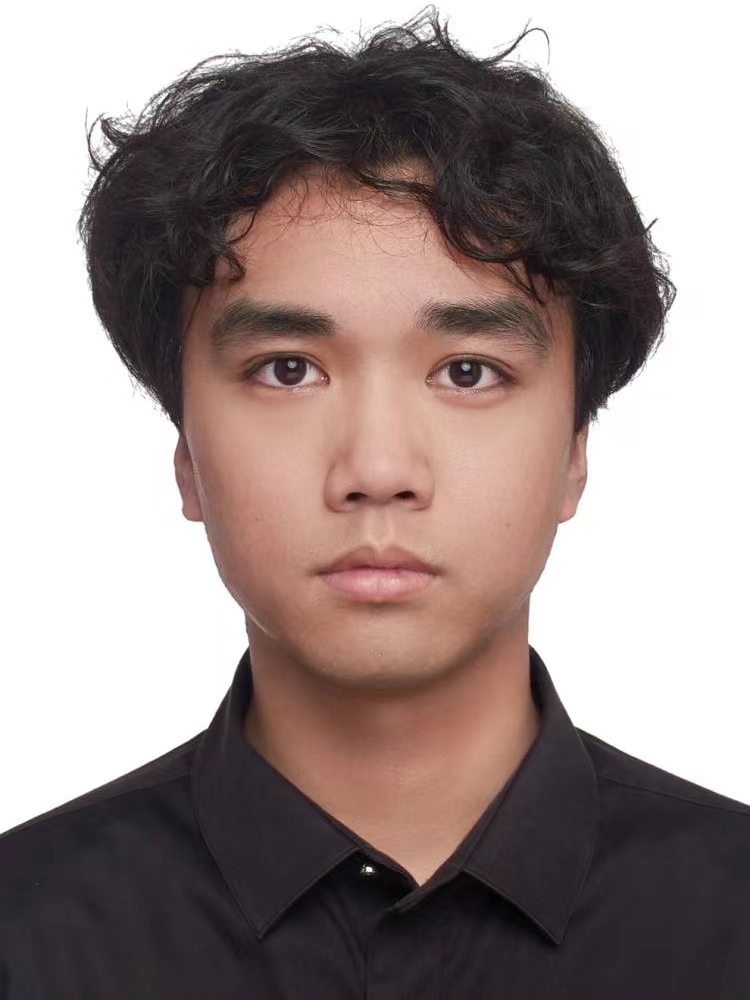}}]{Jie Yang } is currently pursuing a B.S degree in  Artificial Intelligence from  School of Information Science and Chongqing Jiaotong University. His research interests include deep learning, computer vision and pattern recognition.
\end{IEEEbiography}
\begin{IEEEbiography}[{\includegraphics[width=1in,height=1.25in,clip,keepaspectratio]{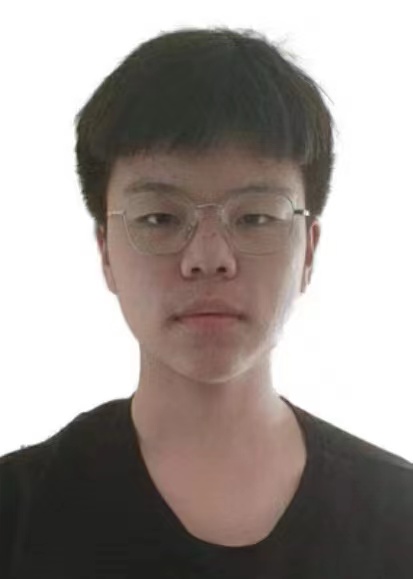}}]{Yudi Ruan} is currently pursuing a B.S degree in  Artificial Intelligence from School of Information Science and Engineering, Chongqing Jiaotong University. His research interests include Few-shot Learning, Domain Adaptation, visual odometry, 3D medical image segmentation, and low-light image enhancement.
\end{IEEEbiography}
\begin{IEEEbiography}[{\includegraphics[width=1in,height=1.25in,clip,keepaspectratio]{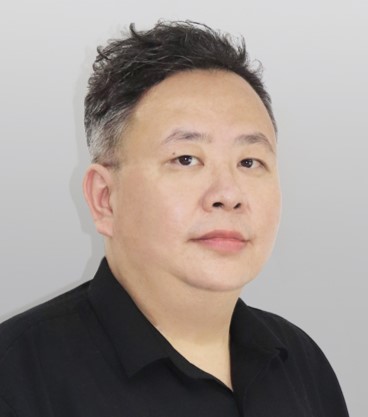}}]{Yuan Li}
 graduated from the College of Engineering at Arizona State University with a master's degree and possesses over 10 years of experience in computer vision and video processing algorithm engineering. He previously worked at Sony, focusing on the research and development of video-related algorithms. Currently, he leads the AI Algorithm Engineering team at Intellicloud's Product and Engineering Research Center, dedicated to the practical application of AI algorithms. 
\end{IEEEbiography}
\begin{IEEEbiography}[{\includegraphics[width=1in,height=1.25in,clip,keepaspectratio]{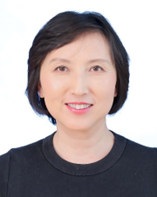}}]{Ying (Gina) Tang}
 received the B.S. and M.S. degrees from the Northeastern University, P. R. China, in 1996 and 1998, respectively, and Ph. D. degree from New Jersey Institute of Technology in 2001. She is currently a Professor of Electrical and Computer Engineering at Rowan University, Glassboro, New Jersey, USA. She is also with the State Key Laboratory for Management and Control of Complex Systems, Institute of Automation, Chinese Academy of Sciences, Beijing, China; and also with the School of Information Science and Technology, Maritime University, Dalian, China. Her current research interests lie in the area of discrete event systems and visualization, including virtual reality/augmented reality, modeling and adaptive control for computer-integrated systems, intelligent game-based learning environments, sustainable production and service automation, blockchain and Petri Nets. 
\end{IEEEbiography}
\vspace{11pt}


\vfill

\end{document}